\documentclass[11pt]{article}
\usepackage[a4paper,margin=24mm]{geometry}
\usepackage[T1]{fontenc}
\usepackage{iftex}
\ifPDFTeX\usepackage[utf8]{inputenc}\fi
\usepackage{lmodern}
\usepackage{amsmath,amssymb,amsthm,booktabs,graphicx,xcolor,array}
\usepackage[numbers,sort&compress]{natbib}
\usepackage{float}
\usepackage[hidelinks]{hyperref}
\usepackage{microtype}
\newcommand{\R}{\mathbb{R}}
\newcommand{\rka}{\mathrm{RKA}}
\newcommand{\softplus}{\operatorname{softplus}}
\newcommand{\sigmoid}{\operatorname{sigmoid}}
\newcommand{\rmsnorm}{\operatorname{RMSNorm}}
\newcommand{\diag}{\operatorname{diag}}
\newcommand{\rank}{\operatorname{rank}}
\newtheorem{proposition}{Proposition}
\title{Radial Kernel Attention for Global Feature Exchange in Polymer Machine Learning Force Fields}
\author{\parbox{\dimexpr\textwidth-2\tabcolsep\relax}{\centering
\normalsize Wenbo Fu\textsuperscript{1}, Chenghui Mao\textsuperscript{1}, Haifeng Lv\textsuperscript{1,3}, Chengwen Hu\textsuperscript{1,*}, Zhou Lu\textsuperscript{1,2,\textdagger}\\[0.6em]
\small \textsuperscript{1}Rhinovate Technology Co., Ltd., Shenzhen, China\\[0.3em]
\textsuperscript{2}Department of Computer Science and Engineering, The Chinese University of Hong~Kong, Hong~Kong, China\\[0.3em]
\textsuperscript{3}School of Materials Science and Engineering, Peking University, Beijing, China\\[0.6em]
\textsuperscript{*}\href{mailto:chengwen@rhinovate.tech}{chengwen@rhinovate.tech}\qquad
\textsuperscript{\textdagger}\href{mailto:zlyu25@cse.cuhk.edu.hk}{zlyu25@cse.cuhk.edu.hk}}}
\date{}
\begin{document}
\maketitle
\begin{abstract}
We introduce Radial Kernel Attention (RKA) for global feature exchange in polymer machine learning force fields. Local neighbor graphs constrain communication across polymer chains and their surrounding environments, while explicit all-pair attention incurs quadratic cost. RKA maps atom-to-centroid distances to learned radial features and combines their overlap with content-dependent queries and keys. Factorized aggregation and readout enable global feature exchange at linear cost in atom count for fixed feature dimensions. We insert the invariant scalar update between EquiformerV3 local blocks and obtain conservative forces by differentiating the total energy, including its radial coordinate dependence. We evaluate energy and force prediction for finite polymer-derived clusters in OPoly26. Our models are trained on the official training split for 12 epochs and evaluated directly after training on the complete official validation split. A matched 5M comparison shows that RKA reduces per-atom energy and force MAEs by 67.4\% and 36.3\%, respectively. The 20M RKA model attains a total-energy MAE of 15.81 meV, a per-atom energy MAE of 0.1182 meV/atom, and a force MAE of 2.619 meV/\AA, improving all three metrics over UMA-S-1.2 and UMA-S-1.2.1 on the same validation structures. RKA provides a compact pathway for information exchange beyond local graph connectivity in polymer machine learning force fields.
\end{abstract}
\section{Introduction}
Representing information beyond a local neighbor graph is an important challenge for atomistic models of polymers. Changes in chain conformation alter bonded geometry, bring distant chain segments into contact, and reorganize interactions with surrounding chains, solvent molecules, and ions~\cite{polymer_roadmap}. A machine learning force field (MLFF) learns energies and forces from electronic-structure data and must remain accurate across these environments~\cite{mlff_review}. Polymer-derived clusters provide a concrete setting in which local chemical structure and the spatial arrangement of the wider system must be represented together. OPoly26 supplies quantum-chemical energy and force data across diverse polymer compositions, chain architectures, and solvation environments, making this setting accessible to learned atomistic models~\cite{opoly26}.

Local geometric networks offer a useful foundation for these models, from distance-based filters to higher-body-order equivariant messages~\cite{schnet,nequip,mace}. They organize computation around nearby atoms and encode how geometric features transform under rotations. EquiformerV3, for example, combines equivariant graph attention with spherical nonlinearities to represent directional local environments~\cite{equiformerv3}. Restricting messages to a distance cutoff controls the number of interactions processed at each layer. The same restriction, however, makes communication depend on the connectivity and depth of the local graph. After a finite number of layers, a node can depend only on nodes within its graph receptive field. Increasing depth extends this field within connected regions, but does not create a communication path between disconnected components.

The spatial extent of polymer clusters makes this distinction relevant. Our analysis of the official OPoly26 training split measures the spatial extent of each of its 5,902,827 structures by the maximum Euclidean distance between any pair of atoms, $D_g=\max_{i,j\in g}\lVert r_i-r_j\rVert_2$. The distribution of $D_g$ across these structures has a median of 23.0~\AA\ and a 90th percentile of 30.7~\AA, compared with the 6~\AA\ cutoff used by our local backbone. These distances characterize spatial extent rather than graph connectivity, but show why a single neighborhood does not describe the whole structure. More fundamentally, a purely local additive model cannot respond to the relative displacement of two disconnected fragments when their internal geometries and all local neighborhoods remain unchanged. Their relative arrangement can still matter physically. A global branch provides a route for such geometric changes to influence the learned representation without requiring a chain of local messages.

Designing this branch requires both a computational choice and a geometric one. Explicit all-pair attention connects every node pair but incurs quadratic work and storage when its interaction matrix is constructed. Factorized attention changes the order of aggregation and querying to avoid that matrix~\cite{linearattention}. For geometric systems, the factorization must also carry information about spatial arrangement while respecting the host model's symmetries. Existing approaches include Fourier-space nonlocal messages, Euclidean positional representations, and all-to-all node attention~\cite{ewaldmp,efa,allscaip}. The question addressed here is whether a compact, structure-relative radial representation can supply a useful global interaction mechanism alongside directional local features.

We introduce Radial Kernel Attention (RKA) for global feature exchange in polymer MLFFs. RKA uses distances from the cluster centroid to organize global feature exchange. Learned, overlapping Gaussian features map each atom into a shared radial representation. The overlap between two radial feature vectors defines a geometric kernel that modulates the content-dependent query--key interaction. RKA accumulates key--value matrices for each radial feature over the entire cluster, then lets each atom read these matrices through its own query and radial weights. Aggregation and readout are linear in atom count for fixed feature dimensions, and no all-pair interaction matrix is formed. The output is a gated scalar residual that can be inserted between local equivariant layers.

The local and radial representations have complementary roles. Local layers encode chemical and directional environments; the radial branch relates those environments to the cluster as a whole. Equal centroid radii produce equal geometric weights, but atoms can still interact differently through their learned queries and keys. Updating scalar channels preserves their geometric invariance and allows later local layers to use the globally updated features. Here, global feature exchange denotes communication across the entire finite structure, including atoms outside a local graph receptive field. Our contribution is the centroid-based radial mechanism and its integration with local equivariant features; its empirical scope is energy and force prediction for finite polymer-derived clusters.

We instantiate RKA in a conservative EquiformerV3 potential by placing a radial branch after each local block. A scalar energy readout defines the total energy, and forces follow from its coordinate derivative. The centroid and radial features remain in the derivative graph, so force supervision trains the global pathway together with the local backbone. Charge/spin conditioning and screened short-range repulsion are shared by the 5M baseline and its RKA counterpart.

We evaluate two complementary comparisons on all 201,865 structures in the official OPoly26 validation split. In the shared 5M architecture, adding RKA reduces per-atom energy MAE by 67.4\% and force MAE by 36.3\%. The complete 20M RKA model improves all three reported MAEs over UMA-S-1.2 and UMA-S-1.2.1~\cite{uma}. Our models use the final parameters after 12 training epochs on the official training split; all models are evaluated on the complete validation split, with no skipped structures. The official test set is unavailable to us. The 5M comparison tests the added pathway within a shared backbone, whereas the UMA comparison assesses complete-model accuracy without matching training budgets.

\begin{samepage}
The contributions of this work are:
\begin{itemize}
    \item A centroid-based radial kernel attention mechanism for global feature exchange in polymer force fields, combining spatial weights with content-dependent attention without constructing an all-pair interaction matrix.
    \item A factorized implementation with linear complexity in atom count at fixed feature dimensions, geometric invariance of the scalar update, and integration with local equivariant layers and conservative force prediction.
    \item An evaluation on OPoly26 that separates the effect of adding RKA to a shared 5M backbone from the predictive performance of a complete 20M RKA model relative to UMA, using official validation for both comparisons.
\end{itemize}
\end{samepage}

\section{Related Work and Problem Formulation}
\label{sec:related}
\paragraph{Local invariant and equivariant representations.}
Distance-based continuous-filter networks such as SchNet encode atomic environments without discretizing positions onto a spatial grid~\cite{schnet}. Equivariant models retain features that transform with rotations: NequIP uses equivariant convolutions, while MACE constructs higher-body-order messages~\cite{nequip,mace}. These approaches enrich local environment descriptions in different ways. The Equiformer family brings attention to equivariant atomic graphs~\cite{equiformer,equiformerv2}. Edge-aligned SO(2) operations reduce the cost of higher-degree representations~\cite{escn}, and EquiformerV3 combines local attention with smooth cutoff weighting and spherical nonlinearities~\cite{equiformerv3}. We use EquiformerV3 as the shared backbone. RKA supplies a scalar communication pathway across the cluster while retaining this directional local processing.

\paragraph{Global attention and geometric kernels.}
Linear attention exploits a factorized similarity function to aggregate features before querying them at individual nodes~\cite{linearattention}. Global attention has also been used in force fields: SpookyNet combines electronic-state inputs with nonlocal attention and analytical energy corrections~\cite{spookynet}, and AllScAIP investigates all-to-all node attention for long-range prediction in energy-conserving MLIPs~\cite{allscaip}. Thus, global attention and conservative forces are established design ingredients. RKA focuses on a specific geometric representation and its factorized evaluation. Euclidean Fast Attention encodes spatial information through Euclidean rotary positional encodings and angular integration~\cite{efa}; RKA instead uses overlap between radial feature maps measured from a common centroid. Its geometric kernel is generally not a function only of interatomic distance. The methodological distinction lies in this spatial representation and its coupling to local features.

\paragraph{Long-range energies and nonlocal messages.}
An alternative route introduces an explicit long-range energy term. DPLR models electrostatics with Gaussian charge distributions at ionic and learned electronic sites and combines this contribution with a short-range potential~\cite{dplr}. Latent Ewald summation learns variables from local descriptors and applies Ewald summation to capture long-range contributions~\cite{les}. Ewald message passing instead supplies nonlocal Fourier-space messages to molecular graph models~\cite{ewaldmp}. RKA supplies nonlocal messages through a finite-dimensional radial feature representation, complementing approaches that explicitly parameterize a long-range energy term.

\paragraph{Setting.}
Let structure $g$ contain $N_g$ nodes, with positions $r_i\in\R^3$ and scalar features $h_i\in\R^C$. A batch consists of disjoint structures, and all sums below are restricted to nodes in the same structure. For geometric invariance, the input features must be invariant under a common rotation and translation of their structure. Relabeling the nodes must correspondingly relabel their input features. RKA returns $h_i'\in\R^C$, allowing subsequent local processing or a task-dependent readout. Atomic identities and electronic-state variables enter through the host model and are not required by the definition of RKA.

Consider a purely local additive energy model evaluated on two fragments with fixed internal geometries. Once every cross-fragment distance exceeds the cutoff, and no other geometry-dependent global operation is present, translating one fragment leaves all local inputs unchanged. The predicted energy is then independent of fragment separation. RKA introduces a route by which changing their arrangement can change the shared radial representation and hence the predicted output. The existence of this route does not, by itself, specify the quantitative distance dependence learned by the model.

\section{Radial Kernel Attention}
\label{sec:method}
\begin{figure}[H]
    \centering
    \includegraphics[width=\linewidth]{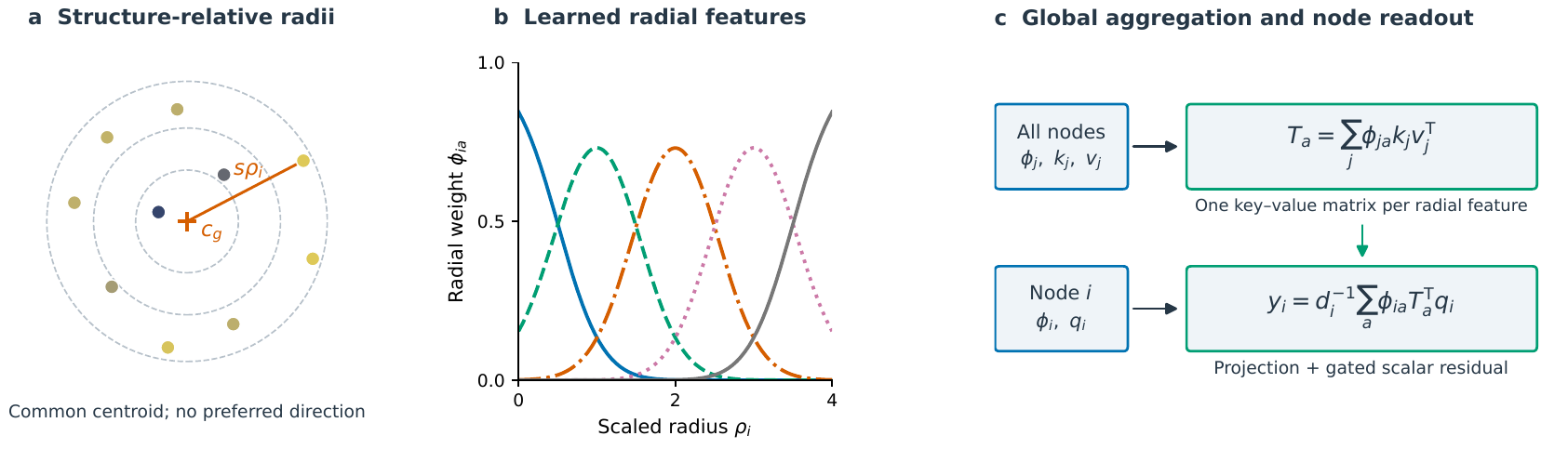}
    \caption{Radial Kernel Attention. A finite structure defines a centroid and node radii. Learned radial features organize content-dependent key--value statistics over the entire structure. Each node reads the shared statistics through its radial features and query, and receives a scalar residual update. The basis curves are schematic, rather than fitted checkpoint values.}
    \label{fig:overview}
\end{figure}

We define RKA for finite point sets with invariant scalar features and integrate it into a polymer MLFF. The centroid provides a shared reference that transforms with the structure and can be computed in linear time. Distances from this reference give rotation- and translation-invariant spatial features. Representing their geometric similarity as a feature inner product makes the interaction separable, so global key--value statistics can be accumulated before individual nodes query them. This radial representation complements the chemical and directional information supplied by the local equivariant backbone. Figure~\ref{fig:overview} summarizes the resulting computation.

\subsection{Centroid-Based Radial Features}
The centroid and scaled radius of each node are
\begin{equation}
c_g=\frac{1}{N_g}\sum_{j\in g}r_j,
\qquad \rho_i=\frac{\lVert r_i-c_g\rVert_2}{s},
\label{eq:radius}
\end{equation}
where $s>0$ is a fixed length scale. The centroid supplies a common reference without choosing a preferred direction. We use $M$ learned Gaussian features,
\begin{equation}
\psi_{ia}=\exp\!\left[-\softplus(\eta_a)(\rho_i-\mu_a)^2\right],
\qquad
\phi_{ia}=\frac{\psi_{ia}}{\max\!\left(\sum_{a'=1}^{M}\psi_{ia'},\epsilon\right)}.
\label{eq:radial}
\end{equation}
The parameters $\mu_a$ and $\eta_a$ determine the radial locations and widths. These are centers on a one-dimensional radius axis, not learned centers in three-dimensional space. Overlapping basis functions let neighboring radial regions share information. The denominator floor avoids division by very small sums; away from that floor, the radial weights sum to one.

The geometric interaction between two nodes is
\begin{equation}
K_{\mathrm{rad}}(i,j)=\sum_{a=1}^{M}\phi_{ia}\phi_{ja}.
\label{eq:kernel}
\end{equation}
This kernel measures radial-feature overlap. Nodes with identical radii have identical $\phi_i$, even when their mutual distances differ. Content-dependent features distinguish their local environments in the attention mechanism below. Because the centroid depends on all nodes, the kernel for a pair can also change when the surrounding structure changes.

\subsection{Content-Dependent Aggregation and Readout}
We compute queries and keys from scalar features normalized by root mean square layer normalization (RMSNorm)~\cite{rmsnorm}, while the values retain the amplitude of the unnormalized inputs:
\begin{equation}
\bar h_i=\rmsnorm(h_i),\qquad
q_i=\softplus(W_q\bar h_i),\qquad
k_i=\softplus(W_k\bar h_i),\qquad
v_i=W_v h_i.
\label{eq:qkv}
\end{equation}
Here $q_i,k_i\in\R^{D_q}$ and $v_i\in\R^{D_v}$. Softplus makes the query and key entries positive. The value projection remains signed and is not constrained to represent a charge or another physical quantity.

For each structure and radial feature, we aggregate a key--value matrix,
\begin{equation}
T_{ga}=\sum_{j\in g}\phi_{ja}k_jv_j^{\mathsf T}
\in\R^{D_q\times D_v}.
\label{eq:aggregate}
\end{equation}
Each node reads these matrices using its own query and radial features:
\begin{equation}
y_i=\frac{1}{d_i}\sum_{a=1}^{M}\phi_{ia}T_{ga}^{\mathsf T}q_i,
\qquad
d_i=\max\!\left(\sum_{a=1}^{M}\sum_{b=1}^{D_q}\phi_{ia}k_{ib},\epsilon\right).
\label{eq:readout}
\end{equation}
The denominator is a node-local scaling factor. It does not sum over sending nodes and does not normalize the rows of the full interaction matrix to one. When the radial features sum to one and the outer floor is inactive, it reduces to $\sum_b k_{ib}$. We therefore interpret RKA as kernelized attention with a local scaling rule, rather than a probability distribution over neighbors.

Expanding Eq.~\eqref{eq:readout} gives the equivalent all-pair form
\begin{equation}
y_i=\sum_{j\in g}A_{ij}v_j,
\qquad
A_{ij}=\frac{q_i^{\mathsf T}k_j}{d_i}K_{\mathrm{rad}}(i,j).
\label{eq:pairwise}
\end{equation}
Equation~\eqref{eq:pairwise} separates the contribution of content similarity from that of radial geometry. The implementation evaluates Eqs.~\eqref{eq:aggregate}--\eqref{eq:readout} and never constructs $A$. Self-interactions are included, and the branch has no pair-distance cutoff.

The output is a gated residual,
\begin{equation}
h_i'=h_i+\gamma\odot W_o y_i,
\qquad \gamma=0.25\sigmoid(\lambda),
\label{eq:residual}
\end{equation}
where $\lambda\in\R^C$ is learned. The output projection is initialized to zero, so the branch initially leaves the host representation unchanged. The gate is initialized to $0.05$ in each channel. Zero initialization concerns the initial residual only; it does not remove the global pathway once the output projection is updated.

\paragraph{Algorithm.}
For a batch of structures, RKA performs the following operations:
\begin{enumerate}
    \item Compute a centroid for each structure and the scaled radius of every node.
    \item Evaluate and normalize the learned Gaussian feature map $\phi$.
    \item Form $q$, $k$, and $v$ from the input scalar features.
    \item Accumulate $T_{ga}$ independently for each structure and radial feature.
    \item Read the statistics at each node, apply $d_i$, and project the result.
    \item Add the channel-gated residual to the input scalar features.
\end{enumerate}

\subsection{Geometric Properties and Complexity}
\label{sec:properties}
\begin{proposition}[Geometric invariance and node relabeling]
If the input scalar features are invariant to a common translation and rotation of a structure, its RKA outputs are invariant to the same transformation. If inputs are relabeled by a node permutation, the outputs are relabeled by that permutation.
\end{proposition}
Under a common rigid transformation $r_i'=Rr_i+t$, the centroid becomes $c_g'=Rc_g+t$. Since $\lVert R(r_i-c_g)\rVert_2=\lVert r_i-c_g\rVert_2$, the radii and radial features remain unchanged. Invariant scalar inputs then give unchanged queries, keys, values, and outputs. A node permutation leaves the centroid and structure-wise sums unchanged and permutes all node-specific quantities consistently, proving the stated relabeling property. These properties concern RKA under the input assumptions; the complete model also depends on its backbone and readout.

For fixed $M$, $D_q$, $D_v$, and $C$, the branch scales linearly with the number of nodes. Constructing the centroid and radial features costs $O(N_g M)$, while aggregation and readout cost $O(N_g M D_qD_v)$. The dense projections add $O(N_g C(D_q+D_v))$ work up to constant factors. The structure-level statistics require $O(MD_qD_v)$ storage per structure. The vectorized implementation additionally creates node-level intermediates of size $O(N_gMD_qD_v)$; force training also retains derivative state. Linear scaling therefore does not mean constant training memory.

For fixed inputs, define lifted feature vectors $\widetilde q_i=q_i\otimes\phi_i$ and $\widetilde k_i=k_i\otimes\phi_i$. Then
\begin{equation}
A=D^{-1}\widetilde Q\widetilde K^{\mathsf T},\qquad
D=\diag(d_1,\ldots,d_{N_g}),\qquad
\rank(A)\leq\min(N_g,MD_q).
\label{eq:rank}
\end{equation}
The geometric kernel is a Gram matrix, but $A$ need not be symmetric because queries and keys differ and $D$ acts on only one side. The factorization enables all-node communication without explicitly constructing the all-pair interaction matrix. With $M=D_q=16$, the rank bound is $\min(N_g,256)$ and permits full rank when $N_g\leq256$. The linear computational cost follows from the factorized evaluation order and does not require $A$ to be rank deficient. It is a property of a single mixing operation, not a bound on the expressivity of the entire nonlinear network.

\subsection{Integration with Local Geometric Models}
Let $X_i^{(t)}$ be a geometric node representation before local block $t$, and let $\widetilde X^{(t)}=\mathcal L_t(X^{(t)})$ be its output. In an equivariant model, write its scalar channel as $h_i=\widetilde X_{i,0,:}^{(t)}$. We set
\begin{equation}
X_{i,0,:}^{(t+1)}=\rka_t(h,r,g)_i,
\qquad
X_{i,\ell>0,:}^{(t+1)}=\widetilde X_{i,\ell>0,:}^{(t)}.
\label{eq:integration}
\end{equation}
The RKA parameters are independent across layers in the evaluated models. At the insertion point, only scalar components are updated. Subsequent local layers and normalization can couple the updated scalars to the remaining representation, so Eq.~\eqref{eq:integration} does not imply that higher-degree features are unaffected throughout the network.

\begin{figure}[H]
    \centering
    \includegraphics[width=\linewidth]{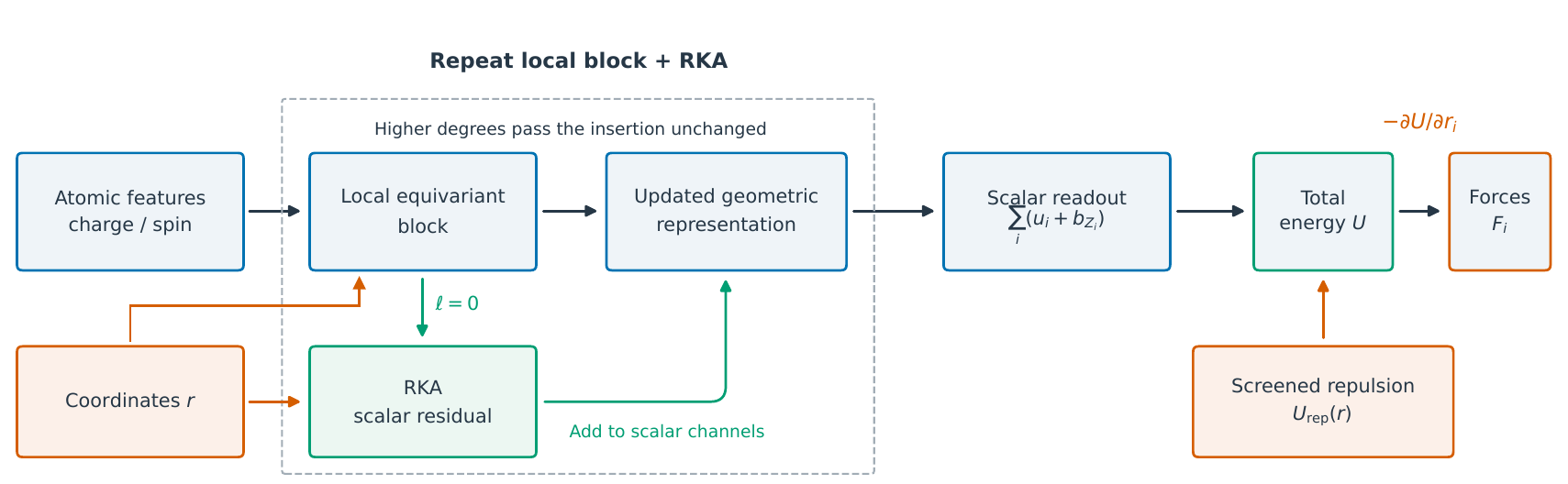}
    \caption{RKA in an equivariant atomic potential. Each local block updates the geometric representation. RKA reads its scalar channels and positions, and writes a scalar residual before the next block. The final scalar readout gives atomic energies. Shared elemental references and screened repulsion complete the total energy, whose coordinate derivative defines the forces. Charge/spin conditioning and repulsion are present in both 5M models.}
    \label{fig:integration}
\end{figure}

Figure~\ref{fig:integration} shows how these scalar updates connect the local equivariant blocks to the energy readout and conservative force prediction.

\subsection{Conservative Polymer Force Field}
We use EquiformerV3 local blocks with a 6~\AA\ neighbor cutoff. Atomic embeddings and charge/spin conditioning initialize the node representation. After the final block and normalization, a scalar readout produces $u_i$. The total energy and forces are
\begin{equation}
U_\theta(g)=\sum_{i\in g}(u_i+b_{Z_i})+U_{\mathrm{rep}}(g),
\qquad F_{i,\theta}=-\frac{\partial U_\theta(g)}{\partial r_i},
\label{eq:energy}
\end{equation}
where $b_{Z_i}$ is a fixed elemental reference and $U_{\mathrm{rep}}$ is Ziegler--Biersack--Littmark (ZBL)-style screened short-range repulsion~\cite{zbl1985}. Both terms are shared across the matched 5M comparison. The supervised energy is $e_\theta(g)=U_\theta(g)/N_g$.

The derivative in Eq.~\eqref{eq:energy} includes the coordinate dependence of the centroid, radial features, and local features. No stop-gradient is applied to the centroid or the radial map. At differentiable configurations, an invariant total energy yields equivariant forces.

The RKA calculations use FP32 when the local model uses reduced-precision computation. Conservative force training retains the coordinate derivative graph for parameter backpropagation.

\section{Experiments}
\label{sec:experiments}
\subsection{Dataset and Evaluation Protocol}
We evaluate on OPoly26, a public dataset of density-functional-theory energies and forces for polymer-derived clusters~\cite{opoly26}. Its structures contain up to 360 atoms and span different compositions, chain architectures, polymerization degrees, and solvation environments. Our models train on the official training split of 5,902,827 structures for 12 epochs. We directly use the model obtained at the end of the twelfth epoch, without selecting an intermediate checkpoint. All models, including the UMA comparators, are evaluated on the same 201,865 structures in the official validation split, with no structures skipped. The official test set is unavailable to us, so the results are validation-set evaluations.

The experiments address two questions. First, does adding RKA improve a 5M model when its local backbone, charge/spin conditioning, elemental references, and screened repulsion are retained? Second, how does the complete 20M RKA model compare with UMA-S-1.2 and UMA-S-1.2.1 on the same structures? UMA is an independently developed pretrained model~\cite{uma}; the latter comparison measures predictive accuracy under a common evaluation protocol, rather than performance at a matched training budget.

\paragraph{Metrics.}
We report only mean absolute error (MAE). For an evaluated set $\mathcal V$ of $G$ structures and $N_{\mathrm{tot}}=\sum_g N_g$ atoms, define $\Delta U_g=U_\theta(g)-U^\star(g)$ and $\Delta F_{i\alpha}=F_{i\alpha,\theta}-F_{i\alpha}^\star$. The metrics are
\begin{align}
\operatorname{MAE}_{U}&=\frac{1}{G}\sum_{g\in\mathcal V}|\Delta U_g|,\\
\operatorname{MAE}_{e}&=\frac{1}{G}\sum_{g\in\mathcal V}\left|\frac{\Delta U_g}{N_g}\right|,\\
\operatorname{MAE}_{F}&=\frac{1}{3N_{\mathrm{tot}}}\sum_{g\in\mathcal V}\sum_{i\in g}\sum_{\alpha\in\{x,y,z\}}|\Delta F_{i\alpha}|.
\label{eq:metrics}
\end{align}
Energy and per-atom energy metrics give equal weight to each structure; the latter divides each energy error by its own atom count before averaging. Force MAE averages Cartesian component errors over all atoms. The reported units are meV per structure, meV/atom, and meV/\AA, respectively. Results are point estimates from the final models, without an estimate of variation across training runs.

\subsection{Models and Training}
The 5M baseline and 5M RKA model share a four-layer EquiformerV3 backbone with 128 channels, maximum angular degree two, and a 6~\AA\ neighbor cutoff. Both include charge/spin conditioning and ZBL-style repulsion. RKA adds one scalar branch after each local block. The 20M model uses seven local layers, maximum angular degree three, a wider feed-forward network, and a higher neighbor cap. It is a larger complete model rather than a parameter-matched control for the 5M architecture. Table~\ref{tab:config} lists the configurations and reported parameter counts.

\begin{table}[H]
\centering
\caption{Configurations of our polymer force fields. The 5M pair shares its local architecture, conditioning, and repulsion. Parameter counts are reported in millions to two decimal places.}
\label{tab:config}
\small
\begin{tabular}{lccc}
\toprule
Setting & 5M baseline & 5M + RKA & 20M + RKA \\
\midrule
Local layers & 4 & 4 & 7 \\
Channels & 128 & 128 & 128 \\
Maximum degree $\ell_{\max}$ & 2 & 2 & 3 \\
Maximum order $m_{\max}$ & 2 & 2 & 2 \\
Feed-forward width & 256 & 256 & 512 \\
Attention heads & 4 & 4 & 4 \\
Neighbor cutoff (\AA) & 6 & 6 & 6 \\
Maximum neighbors & 30 & 30 & 300 \\
Charge/spin conditioning & Yes & Yes & Yes \\
Screened repulsion & Yes & Yes & Yes \\
RKA after each local block & No & Yes & Yes \\
Parameters (M) & 5.48 & 5.56 & 20.62 \\
\bottomrule
\end{tabular}
\end{table}

Every radial branch uses $M=D_q=16$, $D_v=32$, $s=10$~\AA, and a denominator floor of $10^{-6}$. Radial centers are initialized uniformly over $[0,4]$ in scaled-radius units. Raw width parameters are initialized to $\log(1.5)$ before softplus. The output projection is zero-initialized, and the channel gates start at 0.05. Each local layer has its own RKA parameters. The attention spherical grid is $8\times8$ for both sizes, while the feed-forward grid is $8\times8$ for 5M and $10\times10$ for 20M.

Training uses AdamW~\cite{adamw}, seed 3407, weight decay $10^{-3}$, and gradient clipping at norm 100. A 100-step warmup increases the learning rate from $3\times10^{-5}$ to $3\times10^{-4}$, followed by cosine decay toward $3\times10^{-6}$. The 5M configuration uses an atom budget of 2,560 per rank and BF16 local computation with FP32 radial computation. The 20M configuration uses an edge budget of 60,000 per rank and FP32 computation. Candidate batches contain at most 64 structures. These size-dependent batching policies accommodate variable graph sizes.

The training loss combines per-atom energy and vector force errors:
\begin{equation}
\mathcal L_g=10\lvert e_\theta(g)-e^\star(g)\rvert
+\frac{5}{N_g}\sum_{i\in g}\lVert F_{i,\theta}-F_i^\star\rVert_2.
\label{eq:loss}
\end{equation}
The objective is averaged over structures on each rank, and gradients are averaged across ranks. Its atomic vector-norm force error differs from the Cartesian-component MAE reported in Eq.~\eqref{eq:metrics}. Forces are obtained by differentiating total energy, retaining the coordinate derivative graph during parameter backpropagation.

\subsection{Effect of RKA in the 5M Model}
\label{sec:5m}
Adding RKA improves all three MAEs in the matched-backbone comparison (Table~\ref{tab:5m}). Per-atom energy MAE decreases from 0.4665 to 0.1521~meV/atom, a 67.4\% reduction. Total-energy MAE decreases from 59.44 to 21.31~meV per structure, a 64.1\% reduction, while force MAE decreases from 6.632 to 4.223~meV/\AA, a 36.3\% reduction. The RKA model has a reported parameter count of 5.56M.

\begin{table}[H]
\centering
\caption{Effect of RKA on the shared 5M backbone. Both models include charge/spin conditioning and screened repulsion, train for 12 epochs, and use the same official validation structures. Lower MAE is better.}
\label{tab:5m}
\begin{tabular}{lrrr}
\toprule
Metric & 5M baseline & 5M + RKA & Reduction \\
\midrule
Per-atom energy (meV/atom) & 0.4665 & \textbf{0.1521} & 67.4\% \\
Total energy (meV/structure) & 59.44 & \textbf{21.31} & 64.1\% \\
Forces (meV/\AA) & 6.632 & \textbf{4.223} & 36.3\% \\
\bottomrule
\end{tabular}
\end{table}

Both total-energy and per-atom energy MAEs decrease, showing improvements under the two energy aggregation conventions. The accompanying decrease in force MAE extends this improvement to coordinate derivatives of the learned energy. With the local backbone and shared physical components retained, the comparison supports adding the RKA pathway within the evaluated model family.

\subsection{Comparison with Universal Model for Atoms (UMA)}
\label{sec:uma}
UMA is a family of universal interatomic potentials developed by Meta FAIR and pretrained jointly on multiple atomistic datasets~\cite{uma}. UMA-S-1.2 and UMA-S-1.2.1 are two publicly released checkpoints that both use a 64-expert mixture-of-linear-experts (MoLE) architecture, with 291.48M total parameters each. Their training data include OPoly26, making them relevant external baselines for a direct accuracy comparison under the same OPoly26 validation protocol.

Table~\ref{tab:uma} compares the 20M RKA model with two UMA versions on the same complete validation set. The 20M RKA model achieves MAEs of 0.1182~meV/atom for per-atom energy, 15.81~meV per structure for total energy, and 2.619~meV/\AA\ for forces. Both energy MAEs and force MAE are lower than those of either UMA comparator.

\begin{table}[H]
\centering
\caption{Complete-model comparison on all 201,865 official validation structures, with no skipped structures. Parameter counts are the reported model totals in millions. All error columns are MAEs; lower is better.}
\label{tab:uma}
\small
\begin{tabular}{lrrrr}
\toprule
Model & Parameters & Energy/atom & Total energy & Forces \\
 & (M) & (meV/atom) & (meV/structure) & (meV/\AA) \\
\midrule
UMA-S-1.2 & 291.48 & 0.1329 & 17.03 & 3.619 \\
UMA-S-1.2.1 & 291.48 & 0.1258 & 16.09 & 3.616 \\
20M + RKA & 20.62 & \textbf{0.1182} & \textbf{15.81} & \textbf{2.619} \\
\bottomrule
\end{tabular}
\end{table}

Relative to UMA-S-1.2, the reductions are 11.1\% for per-atom energy, 7.2\% for total energy, and 27.6\% for forces. Relative to UMA-S-1.2.1, they are 6.0\%, 1.7\%, and 27.6\%, respectively. The clearest margin is in forces; the total-energy improvement over UMA-S-1.2.1 is smaller.

The complete-model comparison complements the 5M module comparison. It establishes the accuracy attained by the larger RKA potential, but changes in model capacity, training provenance, and architecture prevent attributing its difference from UMA solely to RKA. Reported total parameter counts also do not measure active computation or inference latency. No speed advantage over UMA is inferred from this table.

\section{Discussion}
In the polymer force field, RKA connects atoms throughout a finite cluster through radial statistics and content-dependent readout. This supplies a communication path beyond a local graph's receptive field while keeping the number of geometric channels fixed. The radial map offers a particular geometric inductive bias: atoms interact according to their positions relative to the common centroid, modulated by the chemical and directional environments encoded by the local backbone. The learned energy can consequently depend on remote parts of the cluster even when no sequence of local edges connects them.

The present formulation and evaluation concern finite, nonperiodic polymer-derived clusters. RKA provides structure-level geometric information without specifying an explicit electrostatic or dispersion energy term. The reported results establish predictive gains within the evaluated EquiformerV3 model family on the OPoly26 validation set.

\section{Conclusion}
We introduced Radial Kernel Attention for global feature exchange in polymer machine learning force fields. RKA combines centroid-based radial features with content-dependent attention to exchange information beyond local graph connectivity. Its factorized aggregation and readout have linear cost in atom count for fixed feature dimensions, while its scalar update preserves geometric invariance. Interleaving RKA with local equivariant layers yields a polymer potential whose total energy includes the radial coordinate dependence and defines conservative forces by differentiation. On the shared 5M backbone, RKA reduces per-atom energy MAE by 67.4\% and force MAE by 36.3\%. The complete 20M RKA model improves all three reported MAEs over UMA-S-1.2 and UMA-S-1.2.1 on the same official OPoly26 validation structures. The 20M model achieves reported MAEs of 15.81~meV for total energy and 2.619~meV/\AA\ for forces. RKA offers a compact geometric mechanism for extending the information available to polymer MLFFs, with predictive gains demonstrated on finite polymer-derived clusters.

\section*{Data and Code Availability}
The OPoly26 dataset is publicly available through the official dataset release~\cite{opoly26}. This work uses its official training and validation splits. The code for the RKA models and their training and evaluation will be publicly released in an upcoming update.
\bibliographystyle{unsrtnat}
\bibliography{references}

@misc{equiformerv3,
      title={{EquiformerV3}: Scaling Efficient, Expressive, and General {SE(3)}-Equivariant Graph Attention Transformers}, 
      author={Yi-Lun Liao and Alexander J. Hoffman and Sabrina C. Shen and Alexandre Duval and Sam Walton Norwood and Tess Smidt},
      year={2026},
      eprint={2604.09130},
      archivePrefix={arXiv},
      primaryClass={cs.LG},
      url={https://arxiv.org/abs/2604.09130}, 
}

@misc{efa,
      title={Euclidean Fast Attention -- Machine Learning Global Atomic Representations at Linear Cost}, 
      author={J. Thorben Frank and Stefan Chmiela and Klaus-Robert M{\"u}ller and Oliver T. Unke},
      year={2025},
      eprint={2412.08541},
      archivePrefix={arXiv},
      primaryClass={cs.LG},
      url={https://arxiv.org/abs/2412.08541}, 
}

@misc{opoly26,
      title={The Open Polymers 2026 ({OPoly26}) Dataset and Evaluations}, 
      author={Daniel S. Levine and Nicholas Liesen and Lauren Chua and James Diffenderfer and Helgi Ingolfsson and Matthew P. Kroonblawd and others},
      year={2026},
      eprint={2512.23117},
      archivePrefix={arXiv},
      primaryClass={physics.chem-ph},
      url={https://arxiv.org/abs/2512.23117}, 
}

@misc{uma,
      title={{UMA}: A Family of Universal Models for Atoms}, 
      author={Brandon M. Wood and Misko Dzamba and Xiang Fu and Meng Gao and Muhammed Shuaibi and Luis Barroso-Luque and others},
      year={2026},
      eprint={2506.23971},
      archivePrefix={arXiv},
      primaryClass={cs.LG},
      url={https://arxiv.org/abs/2506.23971}, 
}

@InProceedings{linearattention,
  title = 	 {Transformers are {RNN}s: Fast Autoregressive Transformers with Linear Attention},
  author =       {Katharopoulos, Angelos and Vyas, Apoorv and Pappas, Nikolaos and Fleuret, Fran{\c{c}}ois},
  booktitle = 	 {Proceedings of the 37th International Conference on Machine Learning},
  pages = 	 {5156--5165},
  year = 	 {2020},
  editor = 	 {III, Hal Daumé and Singh, Aarti},
  volume = 	 {119},
  series = 	 {Proceedings of Machine Learning Research},
  month = 	 {13--18 Jul},
  publisher =    {PMLR},
  url = 	 {https://proceedings.mlr.press/v119/katharopoulos20a.html},
}

@InProceedings{ewaldmp,
  title = 	 {Ewald-based Long-Range Message Passing for Molecular Graphs},
  author =       {Kosmala, Arthur and Gasteiger, Johannes and Gao, Nicholas and G\"{u}nnemann, Stephan},
  booktitle = 	 {Proceedings of the 40th International Conference on Machine Learning},
  pages = 	 {17544--17563},
  year = 	 {2023},
  editor = 	 {Krause, Andreas and Brunskill, Emma and Cho, Kyunghyun and Engelhardt, Barbara and Sabato, Sivan and Scarlett, Jonathan},
  volume = 	 {202},
  series = 	 {Proceedings of Machine Learning Research},
  month = 	 {23--29 Jul},
  publisher =    {PMLR},
  url = 	 {https://proceedings.mlr.press/v202/kosmala23a.html},
}

@article{mlff_review, title={Machine Learning Force Fields}, volume={121}, ISSN={1520-6890}, url={http://dx.doi.org/10.1021/acs.chemrev.0c01111}, DOI={10.1021/acs.chemrev.0c01111}, number={16}, journal={Chemical Reviews}, publisher={American Chemical Society (ACS)}, author={Unke, Oliver T. and Chmiela, Stefan and Sauceda, Huziel E. and Gastegger, Michael and Poltavsky, Igor and Schütt, Kristof T. and Tkatchenko, Alexandre and Müller, Klaus-Robert}, year={2021}, month=Mar, pages={10142--10186} }

@article{spookynet, title={{SpookyNet}: Learning force fields with electronic degrees of freedom and nonlocal effects}, volume={12}, ISSN={2041-1723}, url={http://dx.doi.org/10.1038/s41467-021-27504-0}, DOI={10.1038/s41467-021-27504-0}, number={1}, journal={Nature Communications}, publisher={Springer Science and Business Media LLC}, author={Unke, Oliver T. and Chmiela, Stefan and Gastegger, Michael and Schütt, Kristof T. and Sauceda, Huziel E. and Müller, Klaus-Robert}, year={2021}, month=Dec }

@article{dplr, title={A deep potential model with long-range electrostatic interactions}, volume={156}, ISSN={1089-7690}, url={http://dx.doi.org/10.1063/5.0083669}, DOI={10.1063/5.0083669}, number={12}, journal={The Journal of Chemical Physics}, publisher={AIP Publishing}, author={Zhang, Linfeng and Wang, Han and Muniz, Maria Carolina and Panagiotopoulos, Athanassios Z. and Car, Roberto and E, Weinan}, year={2022}, month=Mar }

@misc{schnet,
      title={{SchNet}: A continuous-filter convolutional neural network for modeling quantum interactions}, 
      author={Kristof T. Schütt and Pieter-Jan Kindermans and Huziel E. Sauceda and Stefan Chmiela and Alexandre Tkatchenko and Klaus-Robert Müller},
      year={2017},
      eprint={1706.08566},
      archivePrefix={arXiv},
      primaryClass={stat.ML},
      url={https://arxiv.org/abs/1706.08566}, 
}

@inproceedings{mace,
 author = {Batatia, Ilyes and Kovacs, David P and Simm, Gregor and Ortner, Christoph and Csanyi, Gabor},
 booktitle = {Advances in Neural Information Processing Systems},
 doi = {10.52202/068431-0830},
 editor = {S. Koyejo and S. Mohamed and A. Agarwal and D. Belgrave and K. Cho and A. Oh},
 pages = {11423--11436},
 publisher = {Curran Associates, Inc.},
 title = {{MACE}: Higher Order Equivariant Message Passing Neural Networks for Fast and Accurate Force Fields},
 url = {https://proceedings.neurips.cc/paper_files/paper/2022/file/4a36c3c51af11ed9f34615b81edb5bbc-Paper-Conference.pdf},
 volume = {35},
 year = {2022}
}

@misc{equiformer,
      title={{Equiformer}: Equivariant Graph Attention Transformer for 3D Atomistic Graphs}, 
      author={Yi-Lun Liao and Tess Smidt},
      year={2023},
      eprint={2206.11990},
      archivePrefix={arXiv},
      primaryClass={cs.LG},
      url={https://arxiv.org/abs/2206.11990}, 
}

@misc{equiformerv2,
      title={{EquiformerV2}: Improved Equivariant Transformer for Scaling to Higher-Degree Representations}, 
      author={Yi-Lun Liao and Brandon Wood and Abhishek Das and Tess Smidt},
      year={2024},
      eprint={2306.12059},
      archivePrefix={arXiv},
      primaryClass={cs.LG},
      url={https://arxiv.org/abs/2306.12059}, 
}

@misc{adamw,
      title={Decoupled Weight Decay Regularization}, 
      author={Ilya Loshchilov and Frank Hutter},
      year={2019},
      eprint={1711.05101},
      archivePrefix={arXiv},
      primaryClass={cs.LG},
      url={https://arxiv.org/abs/1711.05101}, 
}

@misc{les,
      title={Latent Ewald summation for machine learning of long-range interactions}, 
      author={Bingqing Cheng},
      year={2024},
      eprint={2408.15165},
      archivePrefix={arXiv},
      primaryClass={cs.LG},
      url={https://arxiv.org/abs/2408.15165}, 
}

@misc{allscaip,
      title={A recipe for scalable attention-based {MLIPs}: unlocking long-range accuracy with all-to-all node attention}, 
      author={Eric Qu and Brandon M. Wood and Aditi S. Krishnapriyan and Zachary W. Ulissi},
      year={2026},
      eprint={2603.06567},
      archivePrefix={arXiv},
      primaryClass={cs.LG},
      url={https://arxiv.org/abs/2603.06567}, 
}

@InProceedings{escn,
  title = 	 {Reducing {SO}(3) Convolutions to {SO}(2) for Efficient Equivariant {GNN}s},
  author =       {Passaro, Saro and Zitnick, C. Lawrence},
  booktitle = 	 {Proceedings of the 40th International Conference on Machine Learning},
  pages = 	 {27420--27438},
  year = 	 {2023},
  editor = 	 {Krause, Andreas and Brunskill, Emma and Cho, Kyunghyun and Engelhardt, Barbara and Sabato, Sivan and Scarlett, Jonathan},
  volume = 	 {202},
  series = 	 {Proceedings of Machine Learning Research},
  month = 	 {23--29 Jul},
  publisher =    {PMLR},
  url = 	 {https://proceedings.mlr.press/v202/passaro23a.html},
}

@article{polymer_roadmap, title={Modeling and Simulations of Polymers: A Roadmap}, volume={52}, ISSN={1520-5835}, url={http://dx.doi.org/10.1021/acs.macromol.8b01836}, DOI={10.1021/acs.macromol.8b01836}, number={3}, journal={Macromolecules}, publisher={American Chemical Society (ACS)}, author={Gartner, Thomas E. and Jayaraman, Arthi}, year={2019}, month=Jan, pages={755--786} }

@article{nequip, title={E(3)-equivariant graph neural networks for data-efficient and accurate interatomic potentials}, volume={13}, ISSN={2041-1723}, url={http://dx.doi.org/10.1038/s41467-022-29939-5}, DOI={10.1038/s41467-022-29939-5}, number={1}, journal={Nature Communications}, publisher={Springer Science and Business Media LLC}, author={Batzner, Simon and Musaelian, Albert and Sun, Lixin and Geiger, Mario and Mailoa, Jonathan P. and Kornbluth, Mordechai and Molinari, Nicola and Smidt, Tess E. and Kozinsky, Boris}, year={2022}, month=May }

@inproceedings{rmsnorm,
 author = {Zhang, Biao and Sennrich, Rico},
 booktitle = {Advances in Neural Information Processing Systems},
 editor = {H. Wallach and H. Larochelle and A. Beygelzimer and F. d\textquotesingle Alch\'{e}-Buc and E. Fox and R. Garnett},
 publisher = {Curran Associates, Inc.},
 title = {Root Mean Square Layer Normalization},
 url = {https://proceedings.neurips.cc/paper_files/paper/2019/file/1e8a19426224ca89e83cef47f1e7f53b-Paper.pdf},
 volume = {32},
 year = {2019}
}

@book{zbl1985,
  title = {The Stopping and Range of Ions in Solids},
  author = {Ziegler, J. F. and Biersack, J. P. and Littmark, U.},
  year = {1985},
  publisher = {Pergamon},
  isbn = {9780080216034},
}
\end{document}